\documentclass{article} % For LaTeX2e
\usepackage{iclr2019_conference,times}

\usepackage{amsmath,amsfonts,bm}

\def\eqref#1{equation~\ref{#1}}
\def\1{\bm{1}}

\DeclareMathAlphabet{\mathsfit}{\encodingdefault}{\sfdefault}{m}{sl}
\SetMathAlphabet{\mathsfit}{bold}{\encodingdefault}{\sfdefault}{bx}{n}

\usepackage{hyperref}
\usepackage{url}
\usepackage{graphicx}
\usepackage{subcaption} 
\usepackage{gensymb}

\title{Visualizing the Consequences of Climate Change Using Cycle-Consistent Adversarial Networks}

\author{Victor Schmidt*, Alexandra Luccioni\thanks{equal contribution}, S. Karthik Mukkavilli,\\ 
\textbf{Kris Sankaran,  \&  Yoshua Bengio}  \\
Montreal Institute for Learning Algorithms\\
Montreal, Canada \\
\texttt{\{schmidtv, luccionis, mukkavis\}@mila.quebec} \\
\And
Narmada Balasooriya  \\
ConscientAI Labs,\\
Colombo, Sri Lanka \\
\And
Jennifer Chayes \\
Microsoft Research New England \\
Cambridge, Massachusetts
}

\iclrfinalcopy % Uncomment for camera-ready version, but NOT for submission.
\begin{document}

\maketitle

\begin{abstract}
We present a project that aims to generate images that depict accurate, vivid, and personalized outcomes of climate change using Cycle-Consistent Adversarial Networks (CycleGANs). By training our CycleGAN model on street-view images of houses before and after extreme weather events (e.g. floods, forest fires, etc.), we learn a mapping that can then be applied to images of locations that have not yet experienced these events. This visual transformation is paired with climate model predictions to assess likelihood and type of climate-related events in the long term (50 years) in order to bring the future closer in the viewer's mind. The eventual goal of our project is to enable individuals to make more informed choices about their climate future by creating a more visceral understanding of the effects of climate change, while maintaining scientific credibility by drawing on climate model projections.
\end{abstract}

\section{Introduction} \label{intro}

It is difficult to downplay the importance of fighting climate change. A recent report from the Intergovernmental Panel on Climate Change has determined that dramatic and rapid changes to the global economy are required in order to avoid increasing climate-related risks for natural and human systems \citep{intergovernmental2018}. However, necessary system overhauls require governmental interventions, which are difficult without strong public support. In fact, recent studies have shown that political will is currently the main obstacle to keeping temperature rise within the limits proposed by the IPCC, i.e. 1.5$^{\circ}$C~\citep{smith2019}.

Unfortunately, public awareness and concern about climate change often does not match the magnitude of its threat to humans and our environment \citep{pidgeon2012,weber2011}. One reason for this mismatch is that it is difficult for people to mentally simulate the complex and probabilistic effects of climate change \citep{oneill2009}.  People often discount the impact that their actions will have on the future, especially if the consequences are long-term, abstract, and at odds with current behavior and identity \citep{espen2016}. To contribute to overcoming these challenges, an easily accessible tool is needed to help the public understand - both rationally and viscerally - the consequences of not taking sufficient action against climate change.

\section{Our Proposal}
\label{proposal}

We propose to develop a Machine Learning (ML) based tool showing in a personalized way the probable effect that climate change will have on a specific location familiar to the viewer. Given an address, it generates an image projecting transformations which are likely to occur there, based on a formal climate model. The hope is that such visualizations would help to visceralize climate change: one might be more willing to take action when seeing the consequences of climate change on their home, their neighbourhood, or the street that they grew up on. The first prototype version of our tool simply generates images of flooded locations based on a binary random variable from a climate model of whether flooding will be present at a given place within a static time frame (in 2050). Eventually, we will extend our model to incorporate other climate-related events (fires and droughts, etc.), varying time horizons, and `decision knobs' allowing the viewer to choose actions and make decisions and see their impact on the projected consequences of climate change. 

In our prototype, we are able to generate images of the projected impact of flooding by training a CycleGAN network~\citep{zhu2017} on Google Street View images of both flooded and unflooded streets and houses~\citep{anguelov2010}. The advantage of using the CycleGAN model is that paired one-to-one mapping is not necessary (i.e. we do not have to have images of the same house before and after a flood). Instead, the model uses domain-level mapping in order to learn the transformation necessary to transform a non-flooded house into a flooded one. We present our approach in more detail in Section~\ref{results}.

\section{Related Work} \label{relatedwork}

Climate change requires solutions to several urgent problems facing humanity and the planet. Since climate sciences ahave entered the era of big data, ML - which has been widely successful in several domains - has brought forward immense potential to contribute to problems in climate sciences. However such applications introduce new challenges for ML due to unique climate physics properties encountered in each problem, requiring novel research in ML. Nonetheless, there are several cross-cutting research themes in problems such as super-resolution, classification, climate down-scaling, forecasting, emulating simulations, localization, detection and tracking of extreme events or anomalies, that are applicable across climate science and ML problems, which requires deep collaboration for synergistic advancements in both disciplines~\citep{monteleoni2013,joppa2017, racah2017, schneider2017, gil2018, hwang2018, karpatne2018, rasp2018}.

Furthermore, ML can help bridge the gap between numerical physics and personalized predictions by improving the accuracy of the physics models. For instance, applying ML techniques along with a physics-guided understanding of meteorology and climate has been shown to significantly improve the prediction of high-impact events~\citep{karpatne2017, rupe2017}. Also, ML techniques can extract otherwise unavailable information from climate forecasts by fusing model output with observations to provide additional decision support for forecasters and users~\citep{mcgovern2017}. Finally, climate science-motivated discovery could lead to advances in ML, as demonstrated in the application of deep learning methods for pixel-level segmentation of extreme events by Kurth et al.~\citeyearpar{kurth2018}.

In this work, we use CycleGANs~\citep{zhu2017} to depict photo-realistic visuals of the potential effects of climate-change events on individual houses and streets. While other approaches to visualize climate change have used both selecting specific images that best represent climate change impacts~\citep{sheppard2012,corner2016} as well as using artistic renderings of possible future landscapes~\citep{giannachi2012} and even video simulations of flooded streets due to rising water levels~\citep{giantasio2014}, to our knowledge, our project is the first application of generative models for the specific purpose of generating images of future climate change impact.

\section{Contributions} \label{results}

While the final version of our visualization tool will include various climate events and incorporate different types of metrics from the climate model, for the initial version of our GAN model, we focused on generating images of houses and buildings specifically after flooding events. In this section, we present the data collection and training approach used for our model.

\subsection{Flooding Image Dataset} \label{data}

One of the most challenging aspects of generating realistic images using GANs is collecting the training data needed in order to extract the mapping function. CycleGAN training assumes that there is some underlying relationship between the domain - for instance a change of seasons in a landscape - which is why we collected many images of streets and houses before and after flooding, with as few extraneous objects (such as vehicles or people) as possible. To collect the necessary training data, we manually searched open source photo-sharing websites for images of houses from various neighborhoods and settings, such as suburban detached houses, urban townhouses, and apartment buildings. We gathered over 500 images of non-flooded houses and the same number of flooded locations, which were all re-sized to 300x300 pixels. 

In order to increase the quantity of images that we could use for training, we performed several data augmentation techniques such as: random crops of a subset of each image, horizontal flipping, small rotations, etc., which enabled us to increase our data set five-fold to over 5000 images total. However, a challenge that we encountered was the fact that flooding is not truly a one-to-one mapping such as the one assumed by the CycleGAN approach, but in fact a many-to-one mapping, i.e. roads, grass, dirt, fences are all mapped to water. For this reason, our data collection was constrained to houses surrounded by lawns, which were then mapped to water by the model. 

 \subsection{Model Architecture and Training}
 
 We use the same architecture for our generative network as that used in the original CycleGAN paper~\citep{he2016}. We trained the networks using the publicly available PyTorch \citep{paszke2017automatic} implementation\footnote{\url{https://github.com/junyanz/pytorch-CycleGAN-and-pix2pix}}. The unique aspect of the CycleGAN approach is the cycle consistency loss, which is used along with the traditional adversarial loss to reduce the space of possible domain-to-domain mapping functions by ensuring that for each image $x$ from domain ${\cal X}$, the image translation cycle should be able to bring $x$ back to the original image (and vice-versa for a given image $y$ from the other domain, ${\cal Y}$).  We trained our CycleGAN model for 200 epochs on the training images, using the Adam solver~\citep{kingma2014} with a batch size of 1, training the model from scratch with a learning rate of 0.0002. As per the CycleGAN training procedure, the learning rate is constant for the first 100 epochs and linearly decayed to zero over the next 100 epochs. We present some of our results below. 
    
%\begin{figure}[h!]
%    \includegraphics[width=\textwidth]{Images/losses.png}
%    \label{lossplot}
%  \caption{CycleGAN training loss curves; even though losses are not known to be very informative for the training of GANs, we can still see from Figure 2 that neither generator has converged, suggesting more work is needed to explore hyper-parameters, models and gather more data}
%\end{figure}

 \subsection{Results}
 
 As can be seen in Figure 1, our CycleGAN model was able to learn an adequate mapping between grass and water, and this mapping could be applied to generate fairly realistic images of flooded houses. The mapping works best with single-family, suburban-type houses which are surrounded by an expanse of grass. There are still improvements to be made with regards to the color scheme of the generated images and the visual artifacts that remain, as well as the coverage of more types of buildings and houses. From the 80 images in the test set, we found that about 70\% were successfully mapped to realistically flooded houses (see~\ref{future} for more information about image evaluation).

\begin{figure}[h!]
  \begin{minipage}{\textwidth}
  \begin{subfigure}[b]{0.5\textwidth}
    \includegraphics[width=\textwidth]{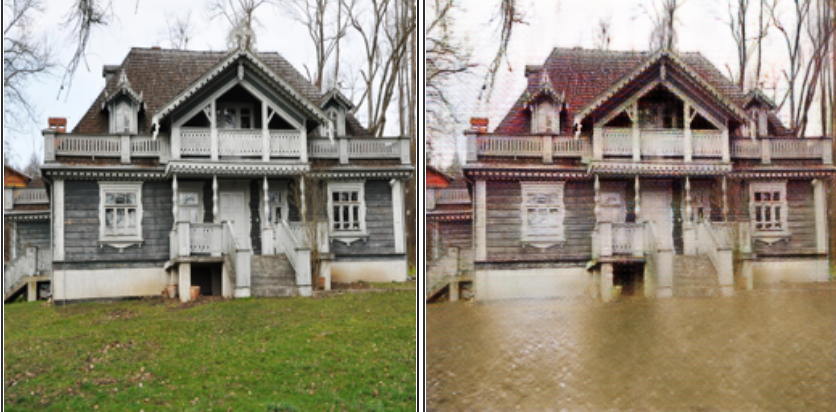}
    \label{fig:1}
  \end{subfigure}
  \begin{subfigure}[b]{0.5\textwidth}
    \includegraphics[width=\textwidth]{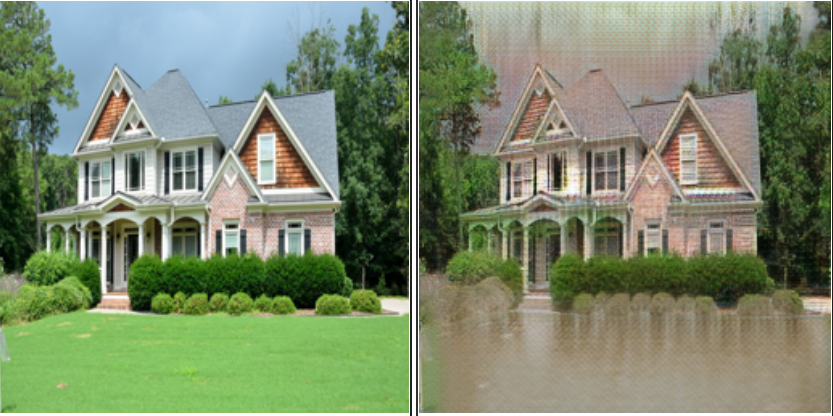}
    \label{fig:2}
  \end{subfigure}
  \caption[Caption for LOF]%
      {Images of flooded houses generated by our model\footnote{Note some artifacts in the sky of the second image, most likely due to trees or clouds in the flooding images used for training}}
    \end{minipage}
\end{figure}

The information about whether or not a house is flooded at specific locations for the CycleGAN images above is sourced from climate model flood hazard outputs, which were converted to binary global flood maps. First, for inland lakes and rivers, a binary flood hazard map was based on each of the 10, 20, 50 and 100 year return runs globally at 1km resolution. We show a 50 year return run in Figure 2 using data from Dottori et al.~\citeyearpar{dottori2016}. Secondly, probabilistic projection data of Extreme Sea Levels until the end of the 21st century along the global coastline was extracted (for the 50th quantile) from Vousdoukas et al.~\citeyearpar{vousdoukas2018} in year 2050 to create a second binary map. This second binary map is based on projections of the decade-window, 2050, under a representation concentration pathway of 4.5 \degree{C} global warming scenario with a greater than 20 cm sea level increase exceedance threshold compared to a baseline sea level from 1980-2014.

\begin{figure}[h!]
  \begin{minipage}{\textwidth}
  \begin{subfigure}[b]{0.5\textwidth}
    \includegraphics[width=\textwidth]{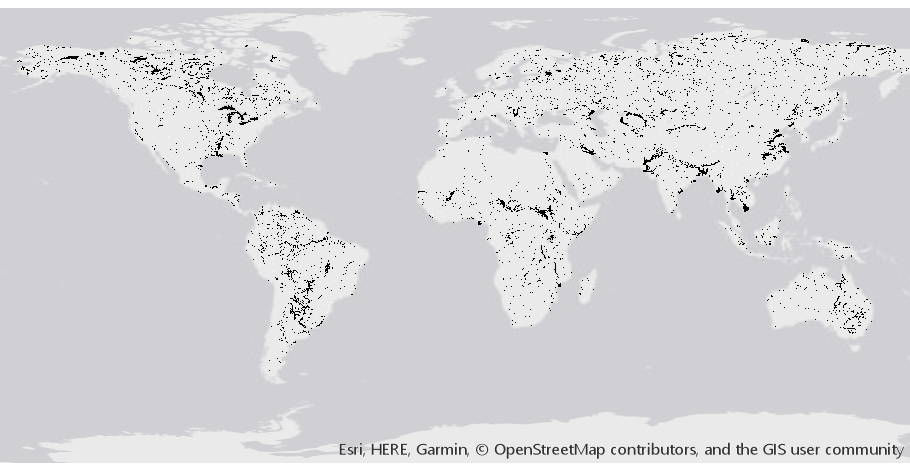}
    \label{floodmap1}
  \end{subfigure}
  \begin{subfigure}[b]{0.5\textwidth}
    \includegraphics[width=\textwidth]{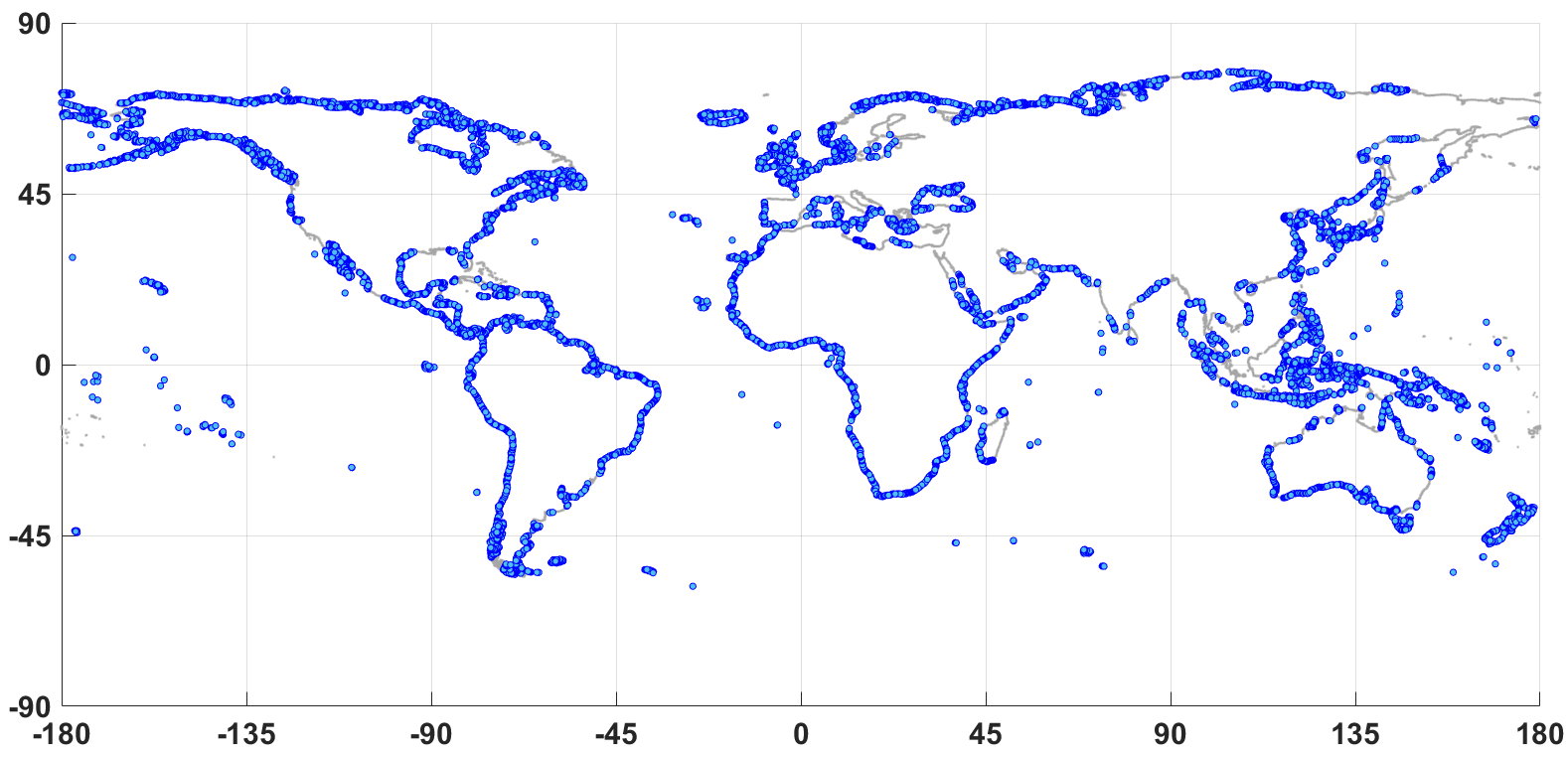}
    \label{floormap2}
  \end{subfigure}
  \caption[Caption for LOF]
      {Binarised maps of global inland lake and river flooding 50-year projection based on Dottori et al.~\citeyearpar{dottori2016} and coastal flood hazard maps for year 2050 based on Vousdoukas et al.~\citeyearpar{vousdoukas2018} \footnote{From left to right: flooding of lakes and rivers inland binary map; binary map of coastal flooding for RCP 4.5 \degree{C} and higher than 20cm rise w.r.t baseline; Colors: black = inland flood hazard, blue = coast flood hazard}}
    \end{minipage}
\end{figure}

\section{Discussion and Future Directions} \label{future}
The initial version of the CycleGAN model that we have developed in the present paper is a prototype to illustrate the feasibility of applying generative models to create personalized images of an extreme climate event, flooding, that is expected to increase in frequency based on climate change projections. Subsequent versions of our model will integrate more varied types of houses and surroundings, as well as different types of climate-change related extreme event phenomena (i.e. droughts, hurricanes, wildfires, air pollution etc), depending on the expected impacts at a given location, as well as forecast time horizons.

Furthermore, to channel the emotional response into behavioural change or actions, another important planned improvement to our model is the eventual addition of `choice knobs', to enable users to visually see the impact of their personal choices, such as deciding to use more public transportation, as well as the impact of broader policy decisions, such as carbon tax and increasing renewable portfolio standards. The effects of an individual turning these knobs could be based on the best available climate model projections, such as the one used for our binary flood map ~\citep{dottori2016}, integrated with economic and policy assessment models. Ultimately, by integrating these `knobs' into our system, we aim to help the general population progress towards greater and more visible public support for climate change migitation steps on a national level, facilitating governmental interventions and helping make the required rapid changes to a global sustainable economy.

However, there are several challenges which we are currently facing that require gaps to be bridged in research at the intersection of climate science and ML. Current climate models make projections based on the physics of fluid motion, energy transfer, mass conservation or chemical transport, not using Deep Learning approaches. Furthermore, the spatial resolutions of these physics models are at best regional, which is much coarser than individual households investigated in this problem. Moreover, the model outputs provide physics variables that are non-trivial to translate into equivalent photo-realistic representations. We therefore believe that there a need to explore physical constraints to GAN training in order to incorporate more physical knowledge into these projections. This is important so that a GAN model will not only transform a house to its projected flooded state, but also take into account the forecast simulations of the flooding event represented by the physical variable outputs and probabilistic scenarios by a climate model for a given location.

%Finally, while many GAN and style transfer application researchers use photo-realism as the method for evaluating model output (e.g.~\cite{li2018, zhu2017}), what we truly want is to produce visceral images to provoke an emotional response, this would require collating surveys with field trials and discussion groups. Recent work in climate change communication has explored ways to allow individuals to engage meaningfully with climate change, approaching it through personal values and experiences as well as visual imagery~\citep{oneill2009,doyle2007}. An extensive, cross-national study carried out the evaluations via discussion groups and online surveys, around different types of climate change images (global, local, catastrophes and solutions), aiming to elucidate different categories of reactions such as ease of understanding, motivation to engage, relatability, salience, and even aesthetics~\citep{chapman2016}. This is the type of evaluation that we propose for our model, collating online survey answers from crowd-sourcing platforms with field trials and discussion groups. Based on the results of these evaluations, and specifically which types of image generation end up having the biggest impact on subjects, we can later improve our model to personalize the images displayed. 

%\subsubsection*{Acknowledgments}
%MSR, Chintan Trivedi, GAN RG
\newpage
\bibliography{ganifyICLR2019}
\bibliographystyle{iclr2019_conference}

\end{document}